# A Joint Pixel and Feature Alignment Framework for Cross-dataset Palmprint Recognition

Huikai Shao, *Student Member IEEE* and Dexing Zhong, *Member IEEE*

*Abstract*—Deep learning-based palmprint recognition algorithms have shown great potential. Most of them are mainly focused on identifying samples from the same dataset. However, they may be not suitable for a more convenient case that the images for training and test are from different datasets, such as collected by embedded terminals and smartphones. Therefore, we propose a novel Joint Pixel and Feature Alignment (JPFA) framework for such cross-dataset palmprint recognition scenarios. Two stage-alignment is applied to obtain adaptive features in source and target datasets. 1) Deep style transfer model is adopted to convert source images into fake images to reduce the dataset gaps and perform data augmentation on pixel level. 2) A new deep domain adaptation model is proposed to extract adaptive features by aligning the dataset-specific distributions of target-source and target-fake pairs on feature level. Adequate experiments are conducted on several benchmarks including constrained and unconstrained palmprint databases. The results demonstrate that our JPFA outperforms other models to achieve the state-of-the-arts. Compared with baseline, the accuracy of cross-dataset identification is improved by up to 28.10% and the Equal Error Rate (EER) of cross-dataset verification is reduced by up to 4.69%. To make our results reproducible, the codes are publicly available at http://gr.xjtu.edu.cn/web/bell/resource.

## I. INTRODUCTION

Biometrics is an efficient and secure authentication technology and has caught more and more attention from the public in recent years [1]. There are many kinds of biometrics emerged, such as fingerprint recognition [2] and face recognition [3], which have been widely used in our life. As one of the popular contactless biometrics, palmprint recognition has the advantages of high security, convenience, and user friendliness [4]. Traditional palmprint recognition is

This work is supported by the National Natural Science Foundation of China (No. 61105021), Natural Science Foundation of Zhejiang Province (No. LGF19F030002), and Natural Science Foundation of Shaanxi Province (No. 2020JM-073).

H. Shao is with the School of Automation Science and Engineering, Xi'an Jiaotong University, Xi'an, Shaanxi 710049, China (e-mail: shaohuikai@stu.xjtu.edu.cn).
D. Zhong is with the School of Automation Science and Engineering, Xi'an Jiaotong University, Xi'an, Shaanxi 710049, China, and State Key Lab. for Novel Software Technology, Nanjing University, Nanjing, 210093, P.R. China (e-mail: bell@xjtu.edu.cn).

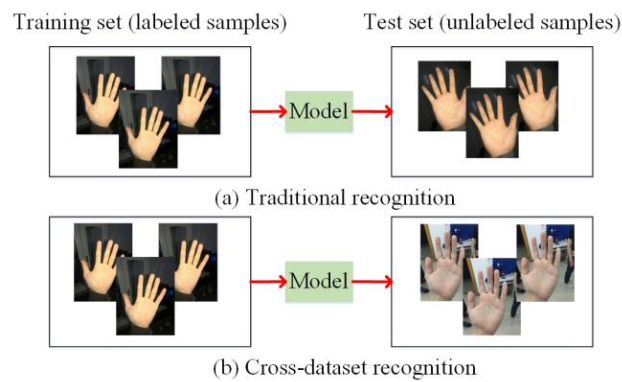

Fig. 1. Illustration of traditional palmprint recognition and cross-dataset recognition. For (a) traditional recognition, the samples of training set and test set are selected from the same database, which are similar in style and illumination. For (b) cross-dataset recognition, the samples of training set and test set come from different databases. Due to the different data distributions, the model trained on the training set cannot be adapted to the test set.

mainly based on the texture features and structural features [5]. A large number of effective algorithms have been applied on specific palmprint databases, such as local discriminant direction binary pattern (LDDBP) [6], PalmNet [7], and apparent and latent direction code (ALDC) [8]. Recently, combined with Convolutional Neural Network (CNN) algorithms, the performance of palmprint recognition has achieved a qualitative breakthrough [9]. Deep learning-based palmprint recognition methods can avoid manual extraction of features to reduce the influence of human factors and significantly improve the performance. For example, Zhong *et al*. [10] implemented an end-to-end palmprint verification system using Deep Hashing Network (DHN) and obtained the state-of-the-arts on benchmarks.

However, most of the current deep learning-based palmprint recognition algorithms have some drawbacks in practical applications. Firstly, most of them are more suitable for identifying samples from the same database, *i.e*., the training set and test set have to be collected by the same environments and devices, which reduces the convenience of palmprint recognition to a certain extent. The general process of palmprint recognition system is that a model is trained based on the training set and then evaluated on the test set. However, due to the dataset gaps, when a model trained on a unique dataset is used directly on another dataset, the performance may become relatively poor [11]. In real-world applications, samples used for training and test may come from different devices, which can be seen as cross-dataset recognition as shown in Fig. 1. For example, we adopt mobile phones to capture palm images for registration and training model, and then terminal devices to capture hand images for authentication. In this case, the model needs to be update based on the later test samples, which will cost time and resources. Secondly, they are mainly supervised algorithms, which requires a large amount of labeled data to guarantee the high accuracy. However, labeling enough data usually needs to take a lot of time, and sometimes it is even unrealistic. For example, Matkowski *et al*. [12] established NTU-PIv1 database and NTU-CP-v1 database. Because the original images are downloaded from the Internet, labeling costs a lot of time and resources.

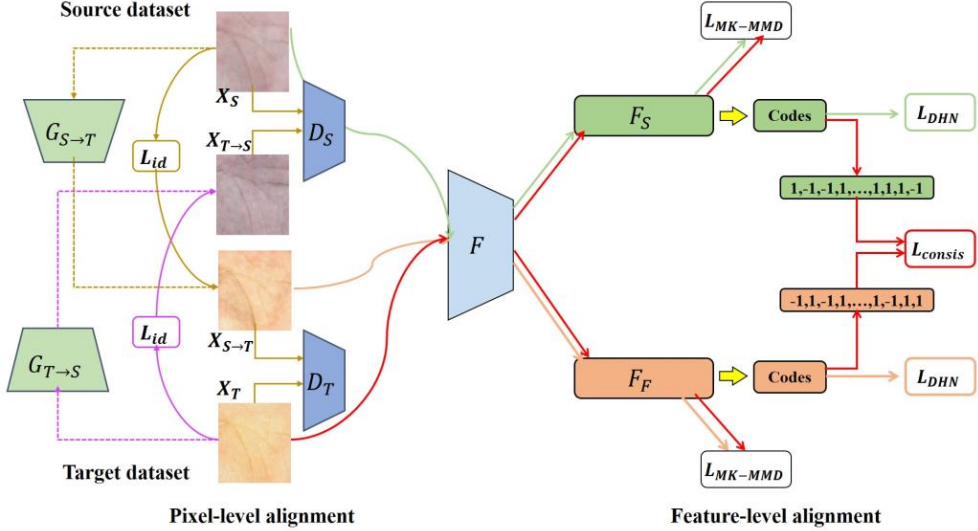

**Fig. 2**. Schematic overview of our framework. On the pixel-level alignment, source palmprint images are converted to labeled fake images similar in style to the target dataset, which can reduce the dataset gap and achieve data augmentation. On the feature-level alignment, a novel deep domain adaptation method is proposed to extracted adaptive features by reducing the distribution difference and consistency loss. (Best viewed in color.)

For cross-dataset palmprint recognition, two datasets collected from different conditions are used as source dataset and target dataset, respectively. The purpose is to match the unlabeled target images with source images to perform identification and verification tasks only using the labels in the source dataset. An effective approach to well address above issue is domain adaptation, which is aimed to use the knowledge learned from source domain to finish the tasks in the target domain [11]. The core of domain adaptation is to extract domain-aligned features in different datasets. Inspired by this idea, in this paper, we proposed a novel Joint Pixel and Feature Alignment (JPFA) framework for cross-dataset palmprint recognition through two-stage alignment, *i.e.*, pixel-level and feature-level alignment, as shown in Fig. 2.

For the first stage, on pixel level, fake images are generated by source images based on style transfer model, which are similar to the images in target dataset in style. It has two advantages, 1) the fake target images can establish the connection between the source dataset and target dataset so that their gap becomes smaller; 2) more labeled images can be obtained for data augmentation to improve the generalization ability of model. The fake images generated need to meet the following conditions. 1) In order to reduce the gap, the fake target images should be as similar as possible to the target images, such as illuminations and textures. 2) In order to obtain labeled data, the fake images should maintain the identity information in the source dataset, *i.e.*, the images generated from the same category should still belong to the same category. The first stage can be achieved based on CycleGAN [13]. Particularly, different from it, a novel identity loss, $L_{id}$, is introduced, which constrains the Euclidean distance between the feature vectors of images before and after transferring.

For the second stage, the features of source and target datasets are aligned to achieve cross-dataset recognition. Firstly, DHN presented in [10] is adopted as feature extractor due to its efficiency of feature matching, but which can be replaced by other models. As shown in Fig. 2, target images are selected randomly to form images pairs with labeled fake and source images. Then image pairs are inputted into a common CNN to get feature maps. Due to the gaps in different datasets, two dataset-specific feature extractors, $F_S$ and $F_F$, are then adopted to obtain specific features, which are also formed as pairs. For every feature pair, Multiple-Kernel Maximum Mean Discrepancy (MK-MMD) is introduce to reduce their distribution differences. In addition, for a palmprint image, the features extracted by $F_S$ and $F_F$ should be as similar as possible. Therefore, a consistency loss $L_{consis}$ is constructed to improve the performance. The details of JPFA can be found in Second III.

The contributions can be summarized as follows:

(1) A novel JPFA framework is proposed for cross-dataset palmprint recognition task through two-stage alignment. Through pixel-level alignment and feature-level alignment, adaptive features can be obtained in source and target datasets. As a result, the unlabeled target images can be matched with source or fake images to find their categories accurately.

(2) On the pixel level, fake images, which are similar to target images in style, are generated to reduce the gap between source and target datasets and perform data augmentation. On the feature level, a new deep domain adaptation model is proposed to align the specific distributions of target-source pair and target-fake pair in the feature spaces

(3) Adequate experiments including cross-dataset palmprint identification and verification on several benchmark databases demonstrate the effectiveness of proposed algorithms. Compare with baseline model, our JPFA can effectively improve accuracy by up to 28.10% and reduce the Equal Error Rate (EER) by up to 4.69%.

Compared with our previous work, PalmGAN, in [14], we have made many significant improvements. Firstly, a new two-stage alignment framework is proposed for cross-dataset palmprint recognition instead of only pixel-level alignment. From the experimental results, JPFA can obtain much better performance than PalmGAN. Secondly, a new task, cross-dataset palmprint identification, is performed to further evaluate the effectiveness of our modified algorithm. Cross-dataset palmprint identification is aimed to match target images with source images, which is more difficult but suitable to practical application. Thirdly, eight additional unconstrained palmprint datasets and two popular benchmarks are also adopted. Fourthly, more analyses and comparisons with the state-of-the-art algorithms including deep and non-deep palmprint recognition methods and domain adaptation methods are provided to further demonstrate the superiority of our algorithms.

The paper consists of 6 sections. Section 2 presents some related works. Section 3 describes our methods in detail. Section 4 shows our experiments and results. Analysis of results in several aspects are presented in section 5. Section 6 concludes the paper.

## II. RELATED WORK

### A. Palmprint Recognition

Palmprint recognition mainly consists of image acquisition, preprocessing, feature extraction and matching [15]. There are many effective algorithms for feature extraction, including statistical, subspace, and coding-based approaches [16]. Wu *et al*. [17] extracted principal lines and wrinkles for authentication. Based on the winner-take-all rule, Jia *et al*. [18] proposed robust line orientation code (RLOC) to extract principal orientation code. Guo *et al*. [19] proposed BOCV representation using Gabor filters on six orientations. Then, Zhang *et al*. [20] further explored BOCV by filtering out the fragile bits and proposed E-BOCV. Zuo *et al*. [21] proposed sparse multiscale competitive code (SMCC) algorithm using a compact representation of multiscale palm line orientation features. Fei *et al*. [22] proposed a discriminative neighboring direction indicator to represent the orientation feature of palmprint. Using a more accurate dominant orientation representation, Xu *et al*. [23] proposed a discriminative and robust competitive code based method for palmprint authentication. Palma *et al*. [24] propose a novel palmprint verification method based on a dynamical system approach, which is used for principal palm lines matching. Fei *et al*. [25] proposed a discriminant direction binary code (DDBC) learning algorithm to form the discriminant direction binary palmprint descriptor for palmprint recognition and proposed discriminant direction binary palmprint descriptor.

Nowadays, many researchers have proposed deep learning-based palmprint recognition methods and obtain promising performance. In 2019, Zhong *et al*. [10] achieved end-to-end palmprint recognition based on DHN and fused it with dorsal hand vein for multi-biometrics. To achieve the touchless palmprint recognition with high-recognition accuracy, Genovese *et al*. [7] proposed PalmNet using a unique method to tune palmprint-specific filters based on Gabor responses and principal component analysis (PCA). Shao *et al*. [26] trained several simple model and combined them together as an ensemble model, call deep ensemble hashing (DEH). Based on generative adversarial network (GAN), Chen *et al*. [27] proposed an effective denoising model for low-resolution palmprint image recognition.

These methods above mentioned mainly carry out palmprint recognition in a single dataset, and basically do not involve the special scenario of cross-dataset recognition, which is more difficult. Due to the dataset gap, they may be not suitable for cross-dataset palmprint recognition and the performances are relatively poor. In [28], Jia *et al*. adopted some traditional code-based methods for palmprint recognition across different devices on three datasets. Ungureanu *et al*. [29] published an unconstrained palmprint database collected by several smartphones and also tried cross-dataset experiments using some classical algorithms. Different from them, in this paper, we propose JPFA based on domain adaptation, which is more suitable and effective to solve such a cross-dataset palmprint recognition issue.

### B. Domain Adaptation

Another line of related work is domain adaptation, which takes the knowledge learned in one domain into another different but relevant domain to perform domain adaptation and finish the target tasks [30, 31]. Therefore, the theory of domain adaptation can be introduced to carry out cross-dataset palmprint recognition. The domain adaptation methods can be divided into two categories: pixel-level adaptation and feature-level adaptation.

Pixel-level adaptation methods usually adopt GAN [32] and its extension algorithms to generate fake images similar to target images in style to reduce the domain gaps between source and target domains. Isola *et al*. [33] proposed conditional GAN to learn a mapping for image-to-image translation application. However, it requires paired training data, which limits its application. More recently, CycleGAN [13] introduces consistency loss to learn the image translation with unpaired data. Based on it, many effective pixel-level adaptation methods have emerged. Wei *et al*. [34] proposed Person Transfer Generative Adversarial Network (PTGAN) to relieve the expensive costs of annotating new training samples and bridge the domain gap for person Re-Identification. Deng *et al*. [35] proposed similarity preserving generative adversarial network (SPGAN) which consists of a Siamese network and a CycleGAN to generate suitable images for domain adaptation and achieved competitive re-ID accuracy. Zhong *et al*. [36] proposed a data augmentation approach that smooths the camera style disparities for camera style (CamStyle) adaptation. Dong *et al*. [37] presented an Asymmetric CycleGAN method with U-net-like generators for translating near-infrared (NIR) face into color (RGB) face. Our first-stage alignment is inspired by these efforts, which is an early attempt to use CycleGAN for cross-dataset palmprint recognition. However, unlike CycleGAN, in addition to maintaining semantic information of real and fake images, a new identity loss is introduced to maintain category information, which can improve the performance significantly.

Feature-level adaptation methods are aimed to obtain alignment features in source and target domains. Pan *et al*. [38] proposed Transfer Component Analysis (TCA) to obtain transfer components across domains in a reproducing kernel Hilbert space based on Maximum Mean Discrepancy (MMD). Then, Long *et al*. [39] proposed Joint Distribution Adaptation (JDA) to simultaneously reduce the difference in both the marginal distribution and conditional distribution between different domains. Nowadays, since deep learning dominates many research fields, there are also many deep transfer learning algorithms proposed. Long *et al*. [40] proposed Deep Adaptation Network (DAN) method, where hidden features of task-specific layers were embedded in a reproducing kernel Hilbert space. Li *et al*. [41] proposed Adaptive Batch Normalization (AdaBN) to achieve deep adaptation performance for domain adaptation tasks. Inspired by GAN [32], there have been many adversarial transfer learning methods. Ganin *et al*. [42] proposed Domain-Adversarial Neural Networks (DANN) to enjoin the network layers to obtain a representation which was predictive of the source labels, but unclear about the domain of input source or target sample. Tzeng *et al*. [43] proposed Adversarial Discriminative Domain Adaptation (ADDA) using discriminative modeling, untied weight sharing, and a GAN loss. ADDA is a general framework, and many existing methods can be considered as special cases of it, such as joint domain alignment and discriminative feature learning for unsupervised deep domain adaptation (JDDA) [44], semantic-aware generative adversarial networks for unsupervised domain adaptation (SeUDA) [45], and Adversarial Autoencoder (AAE) [46].

Recently, there are also some algorithms to perform domain adaptation on both pixel and feature levels. Chen *et al*. [47] adopted image-to-image translation methods for data augmentation and constrained the predictions of translated images between domains for semantic segmentation task consistent. Our methods are more similar to CycADA proposed by Hoffman *et al*. [48]. They firstly generated fake image for target domain and then adopted ADDA for feature-level domain adaptation using target and fake images. Because the domain gap between fake domain and target domain is smaller, ADDA based on fake image is more useful. However, our proposed methods are significantly different from it. On the stage of feature-level adaptation, fake dataset, target dataset, and source dataset are all adopted to improve the generalization of model. In addition, from the results, the difficulty in training the adversarial learning used in ADDA may result in its poor performance. However, combined with MK-MMD and consistency loss, JPFA can obtain alignment features more effectively and avoid collapse of discriminator.

III. METHODS

A. *Deep Hashing Network*

DHN converts palmprint images into binary codes, which can improve the efficiency of feature matching. By optimizing the training, the codes between genuine matches will become as similar as possible, and the codes between imposter matches will become as different as possible, so that the distances between different codes can be calculated to determine whether they are from the same individual [49]. The schematic diagram of DHN is shown in Fig. 3.

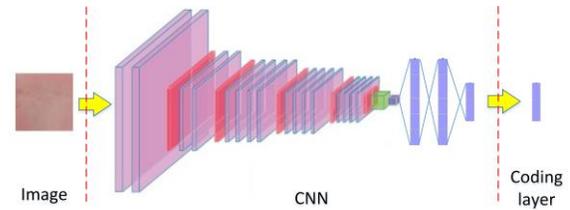

**Fig. 3**. The pipeline of DHN. Palmprint images are inputted into CNN to obtain the discriminative features, which are then converted into hashing codes by coding layer.

Hashing loss and quantization loss between the positive and negative image pairs are adopted to optimize the neural network. Suppose the images $i$ and $j$ as an image pair. The features obtained by CNN are defined as $U_i$ and $U_j$, and their Euclidean distance $D_h(U_i, U_j)$ is adopted to measure their similarity. In order to bring the genuines closer and the imposters farther, hashing loss is defined as

$$L_r(U_i, U_j, S_{ij}) = \frac{1}{2} S_{ij} D_h(U_i, U_j) + \frac{1}{2}(1 - S_{ij}) \max(m - D_h(U_i, U_j), 0), \quad (1)$$

where $m$ is a distance threshold, and $S_{ij}$ is a relationship matrix. If image $i$ and image $j$ come from the same class, $S_{ij} = 1$, otherwise $S_{ij} = 0$.

Further, the loss caused by coding layer is also introduced into the loss function, called quantization loss,

$$L_d(U_i) = \frac{1}{2} \||U_i| - 1\|_2, \quad (2)$$

where $|\cdot|$ is absolute expression and $\|\cdot\|_2$ is $L_2$-norm of vector.

Suppose there are a total of $N$ images, the total loss is

$$L_{DHN} = \sum_{i=1}^{N-1}\sum_{j=i+1}^{N} L_r(U_i, U_j, S_{ij}) + \alpha \sum_{i=1}^{N} L_d(U_i)$$
$$= L_r + \alpha L_d \quad (3)$$

where $\alpha$ controls their importance and is set to 0.5 like [10].

*B. Pixel-level Alignment*

The alignment is firstly performed on pixel level, where fake images similar to target dataset in style are generated using source images. Modified by CycleGAN, for two different image sets $\{x_i \in X\}$ and $\{y_i \in Y\}$ in different domains $X$ and $Y$, CycleGAN maps a sample from source (target) domain to target (source) domain and produces a sample that is indistinguishable from those in the target (source) domain [50]. There are two mapping functions, $G_{X \to Y}$ and $G_{Y \to X}$, and two adversarial discriminators $D_X$ and $D_Y$. The overall CycleGAN loss function is expressed as:

$$L(G_{X \to Y}, G_{Y \to X}, D_X, D_Y) = L_{GAN}(D_Y, G_{X \to Y}, X, Y) + L_{GAN}(D_X, G_{Y \to X}, Y, X) + L_{cyc}(G_{X \to Y}, G_{Y \to X}) \quad (4)$$

where $L_{GAN}$ represents adversarial loss and $L_{cyc}$ represents the cycle consistency loss.

Optimizing $L(G_{X \to Y}, G_{Y \to X}, D_X, D_Y)$ can generate fake palmprint images similar in style to the target dataset. But the fake images should also maintain identities, so identity loss is taken into account. Combined with DHN, if the codes of source and fake images are similar, the category information will be retained. Therefore, identity loss can be achieved by constraining the distances of features. During training, both the source and fake images are inputted into DHN which is pre-trained on source dataset. Then the Euclidean distance between them is calculated as the identity loss,

$$L_{id} = \sum_{i=1}^{N} \|U_i^S - U_i^F\|, \quad (5)$$

where $U_i^S$ and $U_i^F$ are the features of source and fake images extracted by DHN.

So like [14], the optimizing objective on pixel-level alignment can be obtained by

$$L_p = L(G_{X \to Y}, G_{Y \to X}, D_X, D_Y) + L_{id}. \quad (6)$$

*C. Feature-level Alignment*

After the first alignment stage on pixel, there are two labeled datasets that we can use, *i.e.*, source dataset and fake dataset, and the latter is more similar to the target dataset. So in [14], the fake dataset is directly adopted to train DHN in an supervised manner, which is then used for target dataset. However, it does not make full use of labeled data, *i.e.*, more accurate source dataset. Therefore, in this paper, source dataset, fake dataset, and target dataset are all adopted to perform alignment on feature level, as shown in Fig. 2.

Target images are selected to form image pairs with source and fake images, which are inputted into feature extractors to obtain discriminative feature pairs. A common sub-network $F$ is firstly adopted to extract common representations for all images. Then two domain-specific feature extractors, $F_S$ and $F_F$, are applied to obtain binary codes. The main issue in domain adaptation is to reduce the differences of feature distributions between different domains and obtain adaptive alignment features. MMD is often adopted to measure this difference in the Reproducing Kernel Hilbert Space (RKHS). For the source dataset, $X_S$, and target dataset, $X_T$, their MMD loss is

$$L_{MMD} = \left\| \frac{1}{n}\sum_{i=1}^{n}\phi(x_i^S) - \frac{1}{m}\sum_{j=1}^{m}\phi(x_j^T) \right\|_H^2$$
$$= \frac{1}{n^2}\sum_{i=1}^{n}\sum_{i'=1}^{n}\phi(x_i^S)^T\phi(x_{i'}^S) - \frac{2}{mn}\sum_{i=1}^{n}\sum_{j=1}^{m}\phi(x_i^S)^T\phi(x_j^T),$$
$$+ \frac{1}{m^2}\sum_{j=1}^{m}\sum_{j'=1}^{m}\phi(x_j^T)^T\phi(x_{j'}^T) \quad (7)$$

where $n$ and $m$ are the number of samples in the source and target datasets.

MMD applies the kernel trick to measure the squared distance between the empirical kernel mean embeddings. A characteristic kernel $k$ is introduced, which means $k(x_i^S, x_j^T) = \langle \phi(x_i^S), \phi(x_j^T) \rangle$. However, in MMD, the kernel is generally fixed. In order to further improve the performance, Gretton *et al.* [51] proposed MK-MMD, which can find a principled method for optimal kernel selection. So the MK-MMD loss is as follows

$$L_{MK-MMD} = \frac{1}{n^2}\sum_{i=1}^{n}\sum_{i'=1}^{n}k(x_i^S, x_{i'}^S)$$
$$- \frac{2}{mn}\sum_{i=1}^{n}\sum_{j=1}^{m}k(x_i^S, x_j^T) + \frac{1}{m^2}\sum_{j=1}^{m}\sum_{j'=1}^{n}k(x_j^T, x_{j'}^T), \quad (8)$$

and $k$ is defined as the convex combination of $c$ PSD kernels $\{k_u\}$,

$$K \triangleq \left\{ k = \sum_{u=1}^{c}\beta_u k_u : \sum_{u=1}^{c}\beta_u = 1, \beta_u \geq 0, \forall u \right\}. \quad (9)$$

The MK-MMD loss of target-source pairs is defined as $L_{MK-MMD}^{TS}$ and the MK-MMD loss of target-fake pairs is defined as $L_{MK-MMD}^{TF}$.

Further, domain-specific feature extractors are adopted and the target images are transferred into binary codes by them. For the same image, the codes extracted by different extractors should be as similar as possible. Therefore, a consistency loss is introduced by

$$L_{consis} = \left| c_{T_i}^S - c_{T_i}^F \right|, \quad (10)$$

where $c_{T_i}^S$ and $c_{T_i}^F$ are the codes extracted by extractors $F_S$ and $F_F$ for the image $i$ in target dataset.

**Overall objective function**: so the final JPFA can be optimized by combining the losses of two alignment stages on both pixel level and feature level,

$$L = L_{DHN} + L_p + L_{MK-MMD}^{TS} + L_{MK-MMD}^{TF} + \beta L_{consis}, \quad (11)$$

where $\beta$ is a trade-off parameter.

## IV. EXPERIMENTS AND RESULTS

### A. Database

**XJTU-UP (Xi'an Jiaotong University Unconstrained Palmprint) database** [52, 53] is an unconstrained palmprint database collected by mobile phones to promote the practical application of palmprint recognition. Five kinds of mobile phones, *i.e.* iPhone 6S, HUAWEI Mate8, LG G4, Samsung Galaxy Note5, and MI8, are adopted. 100 volunteers held the mobile phones and chose the hand angles and image backgrounds according to their own wishes. Two kinds of illuminations were adopted, one was indoor natural light, and the other was the flash light of smart phones. Each individual provided 200 images of different palms using different mobile phones under different illuminators, respectively. According to different acquisition conditions, XJTU-UP database can be divided into ten different datasets, each of which contains 2,000 palmprint images belonging to 200 categories. For simplicity, according to the acquisition devices and illuminations, ten datasets are denoted as IN (iPhone 6s under Natural illumination), IF (iPhone 6s under Flash illumination), HN (HUAWEI Mate8 under Natural illumination), HF (HUAWEI Mate8 under Flash illumination), LN (LG G4 under Natural illumination), LF (LG G4 under Flash illumination), SN (Samsung Galaxy Note5 under Natural illumination), SF (Samsung Galaxy Note5 under Flash illumination), MN (MI8 under Natural illumination), and MF (MI8 under Flash illumination). In this paper, the images are cropped as region of interests (ROIs) of 224×224 pixels using the method of [53]. Some typical samples are shown in Fig. 4.

**PolyU multispectral palmprint database** is an constrained database acquired under four spectrums, *i.e.* blue, green, red, and near-infrared (NIR) [54]. Volunteers provided several palmprint images in a closed space with a fixed position. According to different illuminations, every independent spectrum can be seen as an independent dataset, *i.e.* NIR, Red, Blue, and Green. In each dataset, there are 6,000 images acquired from 250 individuals between the ages of 20 and 60. Each volunteer was asked to take 12 images for each of left and right hands. All images are cropped to ROIs with a size of 128×128 pixels using the method of [54]. Some examples are shown in Fig. 5.

**Mobile Palmprint Database (MPD)** [55] is collected by two kinds of smartphones, Huawei and Xiaomi, in an unconstrained manner. There are 16,000 palmprint images collected from 200 individuals in two sessions. Each volunteer is asked to provide 10 palm images of each hand in each session using each smartphone. In this paper, we treat the images of left and right hands as different categories. The images collected by different mobile phones are selected as different datasets. So there are two datasets, denoted as HW (Huawei) and Xm (Xiaomi), and each of them contains 8,000 palmprint images belonging to 400 categories. The image is cropped as ROI with a size of 224×224 pixels according to [55]. Fig. 6 shows some typical samples.

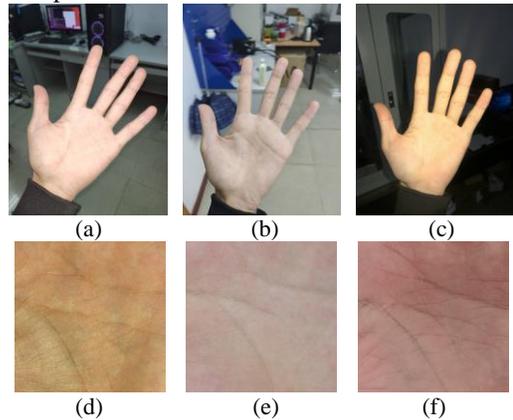

**Fig. 4.** Samples of XJTU-UP database. (a) is in MF, (b) is in MN, (c) is in HF; (d), (e), and (f) are ROIs extracted in LF, LN, and HN, respectively.

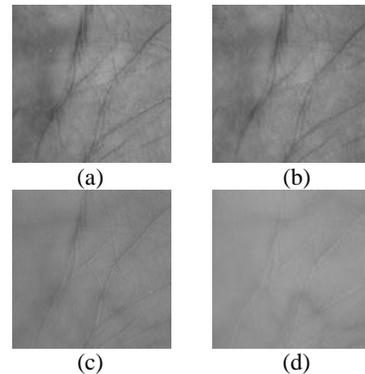

**Fig. 5.** Some typical ROI samples of PolyU multispectral palmprint database. (a) is in Blue, (b) is in Green, (c) is in Red, and (d) is in NIR.

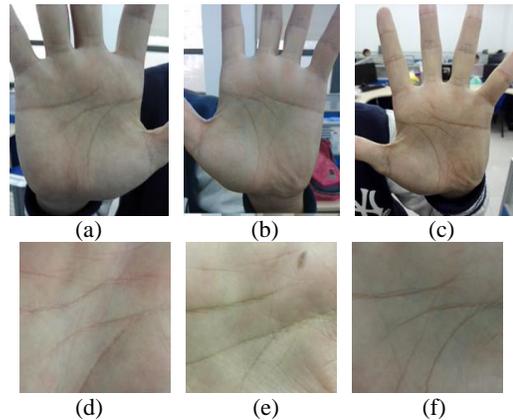

**Fig. 6.** Samples of MPD database. (a) and (c) are in HW, and (b) is in Xm; (d), (e), and (f) are ROIs extracted in HW and Xm.

### B. Implementation Details

In experiments, a dataset is selected as source dataset and another different dataset is selected as target dataset. Each image in the source dataset can generate a fake image similar to other datasets, so a labeled fake dataset with the same number can be obtained. For palmprint identification, every palmprint image in the target dataset is matched with all of the images in the source or fake dataset to find the most similar one. If they are from the same individual, the matching is successful and the identification accuracy can be calculated. For palmprint verification, the hashing codes of target dataset are obtained by the trained feature extractors in the source or fake dataset. After get their Hamming distance, EERs can be calculated. There are two feature extractors which can be used to extract features for target dataset and two labeled datasets which can be used as registration images to match target images. In the experiments, the optimal performance in these two cases are reported as final results. Note that the labels of target dataset are not used in the training progress and only used to evaluate the models.

In the pixel-level alignment, a similar network architecture with the one in CycleGAN [13] is used. In the feature-level alignment, VGG-16 [56] is adopted as the backbone of feature extractor. Specifically, "Batch 1-4" of VGG-16 pre-trained on ImageNet [57] is selected as common feature extractor $F$ and the weights are fixed during training. Then a network architecture like the remaining "Batch 5" and fully connected (FC) layers is adopted as dataset-specific feature extractors $F_S$ and $F_F$. The details are shown in Fig. 7. The experiments are implemented using TensorFlow on a NVIDIA GPU GTX2080 TI with 12G memory power and i9-3.6GHz processors.

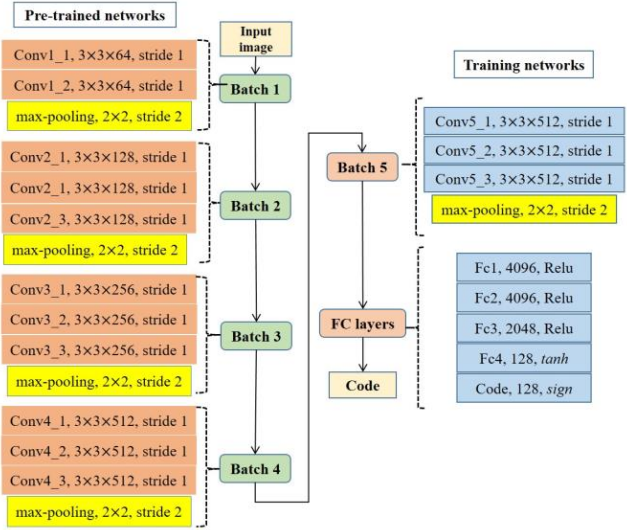

**Fig. 7.** Some details of feature extractor network. For convolutional layers, the parameters of filter size and convolution stride are listed. For max-pooling layers, the windows and strides are given. For FC layers, the dimensions and activation functions are also described.

### C. Palmprint Identification

**Performance on XJTU-UP database**: in XJTU-UP database, SF and IN are selected as source datasets, and the remaining datasets are selected as target datasets. The identification accuracies are listed in Table I. "Source only" is used as a baseline, which means the target codes are obtained by the model trained on source dataset without domain adaptation. From the results, due the various acquisition conditions, the accuracy of different datasets varies greatly. The highest accuracy is 99.00%, where the SF is selected as source dataset and LF is selected as target dataset. The accuracy of cross-dataset identification is improved up to 28.10%.

TABLE I
ACCURACY (%) OF CROSS-DATASET IDENTIFICATION ON XJTU-UP DATABASE

| Source | Target | Source only | JPFA | Improvements |
|---|---|---|---|---|
| SF | LF | 95.30 | **99.00** | ↑ 3.70 |
|  | LN | 73.15 | **82.65** | ↑ 9.50 |
|  | SN | 78.87 | **98.20** | ↑ 19.33 |
|  | IF | 90.10 | **94.85** | ↑ 4.75 |
|  | IN | 60.85 | **70.80** | ↑ 9.95 |
|  | HF | 86.80 | **95.75** | ↑ 8.95 |
|  | HN | 58.35 | **86.45** | ↑ 28.10 |
|  | MF | 94.50 | **96.45** | ↑ 1.95 |
|  | MN | 72.25 | **82.10** | ↑ 9.85 |
| IN | LF | 80.90 | **91.45** | ↑ 10.55 |
|  | LN | 90.35 | **95.20** | ↑ 4.85 |
|  | SF | 79.20 | **85.05** | ↑ 5.85 |
|  | SN | 90.05 | **95.00** | ↑ 4.95 |
|  | IF | 94.20 | **98.10** | ↑ 3.90 |
|  | HF | 67.75 | **88.80** | ↑ 21.05 |
|  | HN | 90.15 | **95.20** | ↑ 5.05 |
|  | MF | 88.45 | **92.05** | ↑ 3.60 |
|  | MN | 91.30 | **95.25** | ↑ 3.95 |

**Performance on PolyU multispectral palmprint database**: Blue and Red are used as source datasets, and the remaining datasets are used as target datasets. The results are shown in Table II. For multispectral palmprint database, the images are collected in constrained environments, so the accuracies are relatively high. When Red and NIR are used as source and target datasets, the improvement is the highest, and the accuracy is improved by 6.45. From the results, when NIR is selected as target dataset, the performance of cross-dataset recognition is poor, which is because the NIR images contain less palmprint information.

TABLE II
ACCURACY (%) OF CROSS-DATASET IDENTIFICATION ON POLYU MULTISPECTRAL PALMPRINT DATABASE

| Source | Target | Source only | JPFA | Improvements |
|---|---|---|---|---|
| Blue | Green | 99.42 | 99.85 | ↑ 0.43 |
|  | Red | 99.80 | 100.00 | ↑ 0.20 |
|  | NIR | 92.57 | 96.43 | ↑ 3.86 |
| Red | Blue | 99.33 | 99.50 | ↑ 0.17 |
|  | Green | 99.27 | 99.53 | ↑ 0.26 |
|  | NIR | 90.52 | 96.97 | ↑ 6.45 |

TABLE III
ACCURACY (%) OF CROSS-DATASET IDENTIFICATION ON MPD

| Source | Target | Source only | JPFA | Improvements |
|---|---|---|---|---|
| HW | Xm | 89.80 | 92.74 | ↑ 2.94 |
| Xm | HW | 93.31 | 94.65 | ↑ 1.34 |

**Performance on MPD**: in MPD, there are two palmprint dataset collected from the same subjects. So the one dataset is selected as source dataset and the other is selected as target dataset. The accuracies are shown in Table III. When HW is used as source dataset and Xm is used as target dataset, the accuracy is improved by 2.94%. When Xm is used as source dataset and HW is used as target dataset, the accuracy is improved by 1.34%.

*D. Palmprint Verification*

**Performance on XJTU-UP database**: palmprint verification is a one-to-one comparison. For XJTU-UP database, SF and IN are still selected as source datasets, and the remaining datasets are selected as target datasets. The EERs of different experiment settings are shown in Table IV. From the results, the performances of cross-dataset palmprint verification are improved significantly. The EER is reduced by up to 4.69%, when SF is used as source dataset and HN is used as target dataset. And the EER of cross-dataset verification is as low as 0.082%, when IN and MN are used as source and target datasets, respectively.

TABLE IV
EER (%) OF CROSS-DATASET VERIFICATION ON XJTU-UP DATABASE

| Source | Target | Source only | JPFA | Improvements |
|---|---|---|---|---|
| SF | LF | 0.74 | **0.16** | ↓ 0.58 |
|  | LN | 5.27 | **2.61** | ↓ 2.66 |
|  | SN | 3.42 | **0.52** | ↓ 2.90 |
|  | IF | 0.95 | **0.30** | ↓ 0.65 |
|  | IN | 4.91 | **1.96** | ↓ 2.95 |
|  | HF | 2.49 | **1.08** | ↓ 1.41 |
|  | HN | 6.78 | **2.09** | ↓ 4.69 |
|  | MF | 1.04 | **0.28** | ↓ 0.76 |
|  | MN | 4.36 | **1.85** | ↓ 2.51 |
| IN | LF | 2.28 | **0.20** | ↓ 2.08 |
|  | LN | 2.00 | **0.85** | ↓ 1.15 |
|  | SF | 2.24 | **0.45** | ↓ 1.79 |
|  | SN | 1.58 | **0.50** | ↓ 1.08 |
|  | IF | 1.12 | **0.33** | ↓ 0.79 |
|  | HF | 4.19 | **1.74** | ↓ 2.45 |
|  | HN | 1.73 | **0.92** | ↓ 0.81 |
|  | MF | 1.76 | **1.14** | ↓ 0.62 |
|  | MN | 1.20 | **0.082** | ↓ 1.12 |

**Performance on PolyU multispectral palmprint database**: the results on PolyU multispectral palmprint database are shown in Table V. When Blue and NIR are selected as source and target datasets, the EER is reduced by up to 2.26%. When NIR is selected as target dataset, the EERs are also the worst. However, for Blue and Green, the EERs of both experiment settings are 0%, due to their similar illuminations.

TABLE V
EER (%) OF CROSS-DATASET VERIFICATION ON POLYU MULTISPECTRAL PALMPRINT DATABASE

| Source | Target | Source only | JPFA | Improvements |
|---|---|---|---|---|
| Blue | Green | 0.17 | 0.055 | ↓ 0.115 |
|  | Red | 0.00 | 0.00 | ↓ 0.00 |
|  | NIR | 4.03 | 1.77 | ↓ 2.26 |
| Red | Blue | 0.26 | 0.073 | ↓ 0.187 |
|  | Green | 0.30 | 0.034 | ↓ 0.266 |
|  | NIR | 0.44 | 0.25 | ↓ 0.19 |

**Performance on MPD**: The EERs of palmprint verification are shown in Table VI. When HW is used as source dataset and Xm is used as target dataset, the EER is reduced by 2.60%. When Xm is used as source dataset and HW is used as target dataset, the EER is reduced by 1.69%. It also shows the efficiency of our methods.

TABLE VI
EER (%) OF CROSS-DATASET VERIFICATION ON MPD

| Source | Target | Source only | JPFA | Improvements |
|---|---|---|---|---|
| HW | Xm | 6.45 | 3.85 | ↓ 2.60 |
| Xm | HW | 4.48 | 2.79 | ↓ 1.69 |

V. EVALUATION AND ANALYSES

*A. Result Analysis*

In this paper, a novel method is proposed for cross-dataset palmprint identification and verification. The results show that it can improve the performance significantly.

**Comparisons between different databases**: in the experiments, three palmprint databases are adopted to evaluate our proposed methods, including sixteen datasets. XJTU-UP database is collected by mobile phones in an unconstrained manner using uneven illumination and uneven hand angles. Multispectral database is collected in enclosed space and the hands are fixed, so the images are least affected by illuminations and noises. MPD is also collected by mobile phones in natural illumination. From the results, the performances of multispectral database are much better than that of XJTU-UP and MPD. Though multispectral database has higher accuracy, it limits the convenience of palmprint recognition, which requires dedicated devices and specific usage. In practical applications, the unconstrained acquisitions are more suitable and flexible, which may be a trend in smart society. In fact, through JPFA, the accuracy and EER of cross-dataset palmprint recognition can be improved very well.

**Comparisons between different illuminations**: for XJTU-UP and MDP, two kinds of illuminations are adopted during acquisition. Flash lights can eliminate the interference of other illuminations, making the images less affected by noise. But in natural illuminations, the environments are complex, so the image collected are difficult to identify. For PolyU multispectral palmprint, when NIR is selected as target dataset, the results are worse, especially when Blue is used as source dataset. NIR images are collected under near-infrared illumination, which is relatively weak. In addition, from the NIR palmprint images, there are many vein patterns, but which contain less information.

*B. Ablation Study*

1) The roles of different losses

In this paper, several effective loss functions are adopted. Here, we conduct experiments to verify the roles that MK-MMD and consistency losses play. SF is selected as source dataset and others are selected as target datasets. Tables VII and VIII show the results. "MK-MMD t-s" means the MK-MMD loss of target-source pairs, and "MK-MMD t-f" means the MK-MMD loss of target-fake pairs. From the results, when MK-MMD loss is adopted on single target-source or target-fake pairs, the performances are similar. When the MK-MMD loss is adopted on both target-source and target-fake pairs, the performances are improved, which demonstrates the usefulness of fake images. However, when consistency loss is also added, the performances are improved significantly on both palmprint identification and verification, which shows the necessity of consistency loss and the effectiveness of our methods.

TABLE VII
ACCURACY (%) OF CROSS-DATASET IDENTIFICATION USING DIFFERENT LOSSES

| Target | MK-MMD (t-s) | MK-MMD (t-f) | MK-MMD (t-s + t-f) | MK-MMD (t-s + t-f) + consistency loss |
|---|---|---|---|---|
| LF | 97.80 | 96.55 | 98.90 | **99.00** |
| LN | 78.45 | 61.05 | 78.05 | **82.65** |
| SN | 90.20 | 96.65 | 97.75 | **98.20** |
| IF | 93.45 | 92.55 | 93.15 | **94.85** |
| IN | 68.45 | **71.75** | 69.60 | 70.80 |
| HF | 93.65 | 93.35 | 95.05 | **95.75** |
| HN | 70.65 | 83.75 | 83.85 | **86.45** |
| MF | 94.95 | 95.40 | 97.20 | **96.45** |
| MN | 80.20 | 71.90 | 81.35 | **82.10** |

TABLE VIII
EER (%) OF CROSS-DATASET VERIFICATION USING DIFFERENT LOSSES

| Target | MK-MMD (t-s) | MK-MMD (t-f) | MK-MMD (t-s + t-f) | MK-MMD (t-s + t-f) + consistency loss |
|---|---|---|---|---|
| LF | 0.58 | 0.20 | 0.20 | **0.16** |
| LN | 2.87 | 3.71 | 3.43 | **2.61** |
| SN | 1.17 | 0.88 | 0.87 | **0.52** |
| IF | 0.35 | 0.47 | 0.32 | **0.30** |
| IN | 2.41 | 2.34 | 2.06 | **1.96** |
| HF | 1.95 | 1.84 | **1.01** | 1.08 |
| HN | 3.99 | **1.94** | 2.43 | 2.09 |
| MF | 0.38 | 0.59 | 0.92 | **0.28** |
| MN | 1.69 | 1.61 | **1.03** | 1.85 |

2) The effect of hyperparameters

The above description explains the effectiveness of consistency loss. Here, several experiments are conducted to analyze the weight of consistency loss, *i.e.*, the effect of hyperparameters $\beta$. SF is also used as source dataset and the others are used as target datasets. The results are shown in Tables IX and X. When $\beta = 1.5$, the performances of both cross-dataset palmprint identification and verification can reach the highest level.

TABLE IX
ACCURACY (%) OF CROSS-DATASET IDENTIFICATION USING DIFFERENT HYPERPARAMETERS

| Target | Source only | $\beta=0.5$ | $\beta=1$ | $\beta=1.5$ | $\beta=2$ | $\beta=2.5$ |
|---|---|---|---|---|---|---|
| LF | 95.30 | **99.20** | 99.00 | 99.00 | 98.20 | 98.65 |
| LN | 73.15 | 81.00 | 81.60 | **82.65** | 75.55 | 78.70 |
| SN | 78.87 | 97.15 | 97.30 | **98.20** | 97.40 | 97.30 |
| IF | 90.10 | 93.00 | 94.30 | **94.85** | 93.40 | 93.85 |
| IN | 60.85 | **74.05** | 69.90 | 70.80 | 73.00 | 73.40 |
| HF | 86.80 | **96.20** | 95.65 | 95.75 | 95.35 | 94.20 |
| HN | 58.35 | 83.60 | 85.10 | **86.45** | 85.30 | 85.20 |
| MF | 94.50 | 95.65 | 95.50 | **96.45** | 95.90 | 96.40 |
| MN | 72.25 | 78.80 | 81.20 | **82.10** | 79.45 | 81.15 |

TABLE X
EER (%) OF CROSS-DATASET VERIFICATION USING DIFFERENT HYPERPARAMETERS

| Target | Source only | $\beta=0.5$ | $\beta=1$ | $\beta=1.5$ | $\beta=2$ | $\beta=2.5$ |
|---|---|---|---|---|---|---|
| LF | 0.74 | 0.20 | 0.22 | **0.16** | 0.19 | 0.20 |
| LN | 5.27 | 3.00 | 3.09 | **2.61** | 3.53 | 3.40 |
| SN | 3.42 | 0.80 | 0.53 | **0.52** | 0.84 | 0.72 |
| IF | 0.95 | 0.33 | 0.38 | **0.30** | 0.60 | 0.49 |
| IN | 4.91 | 2.21 | **1.96** | 1.96 | 2.11 | 2.37 |
| HF | 2.49 | 1.27 | 1.85 | **1.08** | 1.10 | 1.76 |
| HN | 6.78 | 2.33 | 2.24 | **2.09** | 2.53 | 2.34 |
| MF | 1.04 | 0.50 | 0.38 | **0.28** | 0.52 | **0.28** |
| MN | 4.36 | 1.58 | **1.41** | 1.85 | 1.79 | 1.79 |

TABLE XI
COMPARISON RESULTS (ACCURACY, %) OF CROSS-DATASET IDENTIFICATION ON DIFFERENT MODELS

| Source | SF | | | | | | | | |
|---|---|---|---|---|---|---|---|---|---|
| Target | LF | LN | SN | IF | IN | HF | HN | MF | MN |
| DHN | 95.30 | 73.15 | 78.87 | 90.10 | 60.85 | 86.80 | 58.35 | 94.50 | 72.25 |
| ALDC | 92.90 | 66.75 | 76.60 | 83.60 | 58.25 | 87.25 | 64.90 | 88.45 | 64.90 |
| DDBPD | 93.75 | 77.30 | 88.90 | 88.30 | 74.75 | 92.60 | 79.75 | 92.95 | 81.90 |
| LDDBP | 96.60 | 80.85 | 90.75 | 90.80 | 69.05 | 94.00 | 83.55 | 95.05 | 80.70 |
| PalmNet | 76.25 | 61.10 | 65.35 | 74.60 | 53.85 | 77.20 | 56.00 | 74.80 | 57.50 |
| DEH (activation) | 90.24 | 72.15 | 85.33 | 91.20 | 66.32 | 89.20 | 85.65 | 83.94 | 78.32 |
| DEH (adversarial) | 87.35 | 69.94 | 80.95 | 90.03 | 61.44 | 82.13 | 78.85 | 77.34 | 72.10 |
| DAN | 97.80 | 78.45 | 90.20 | 93.45 | 68.45 | 93.65 | 70.65 | 94.95 | 80.20 |
| ADDA | 95.30 | 74.95 | 81.60 | 89.95 | 61.85 | 87.90 | 59.25 | 93.80 | 68.85 |
| Deep CORAL | 65.20 | 57.00 | 63.11 | 83.85 | 56.44 | 57.90 | 40.35 | 89.00 | 58.50 |
| CycleGAN | 73.95 | 64.70 | 60.05 | 61.10 | 69.80 | 63.90 | 69.65 | 70.24 | 67.30 |
| PalmGAN | 95.39 | 67.05 | 93.80 | 90.75 | 68.05 | 86.40 | 79.30 | 93.50 | 66.40 |
| CycADA | 95.35 | 68.30 | 94.20 | 90.73 | 69.03 | 87.40 | 77.30 | 94.04 | 64.57 |
| JPFA (ours) | **99.00** | **82.65** | **98.20** | **94.85** | **70.80** | **95.75** | **86.45** | **96.45** | **82.10** |

TABLE XII
COMPARISON RESULTS (EER, %) OF CROSS-DATASET VERIFICATION ON DIFFERENT MODELS

| Source | SF | | | | | | | | |
|---|---|---|---|---|---|---|---|---|---|
| Target | LF | LN | SN | IF | IN | HF | HN | MF | MN |
| DHN | 0.74 | 5.27 | 3.42 | 0.95 | 4.91 | 2.49 | 6.78 | 1.04 | 4.36 |
| ALDC | 4.09 | 6.02 | 6.93 | 3.85 | 7.29 | 4.22 | 7.43 | 4.07 | 6.59 |
| DDBPD | 3.37 | 4.10 | 4.07 | 2.82 | 4.70 | 3.13 | 4.56 | 3.14 | 3.62 |
| LDDBP | 3.45 | 3.20 | 3.31 | 2.58 | 3.74 | 2.78 | 4.00 | 2.91 | 3.40 |
| PalmNet | 3.24 | 3.48 | 2.80 | 2.82 | 2.76 | 2.45 | 2.32 | 2.97 | 2.55 |
| DEH (activation) | 2.05 | 8.53 | 7.47 | 2.67 | 7.71 | 5.30 | 8.99 | 2.75 | 7.87 |
| DEH (adversarial) | 6.66 | 11.62 | 10.58 | 6.81 | 10.62 | 8.91 | 11.93 | 6.38 | 10.11 |
| DAN | 0.58 | 2.87 | 1.17 | 0.35 | 2.41 | 1.95 | 3.99 | 0.38 | 1.69 |
| ADDA | 0.70 | 4.56 | 3.13 | 1.23 | 5.20 | 2.21 | 5.85 | 1.02 | 3.90 |
| Deep CORAL | 5.22 | 6.02 | 4.32 | 2.17 | 15.16 | 8.90 | 9.94 | 1.55 | 5.75 |
| CycleGAN | 1.55 | 6.35 | 4.48 | 1.56 | 2.23 | 2.46 | 4.60 | 1.24 | 3.98 |
| PalmGAN | 0.83 | 6.01 | 1.88 | 0.98 | 4.16 | 2.58 | 3.58 | 1.03 | 4.51 |
| CycADA | 0.83 | 4.76 | 1.75 | 0.97 | 3.59 | 2.54 | 3.59 | 0.92 | 4.72 |
| **JPFA (ours)** | **0.16** | **2.61** | **0.52** | **0.30** | **1.96** | **1.08** | **2.09** | **0.28** | **1.85** |

## C. Comparisons with Other Models

In order to evaluate the superiority of JPFA for cross-dataset palmprint recognition, we conducted adequate experiments to compare JPFA with the state-of-the-arts including deep and non-deep palmprint recognition methods and domain adaptation methods, as following:

- **DHN** [10] transfers palmprint images into binary codes to improve the efficiency of feature matching. Here, DHN is adopted as feature extractor and a baseline.
- **ALDC** [8] is a novel double-layer direction extraction method. It firstly extracts the apparent direction from the surface layer and exploits the latent direction features from the energy map.
- **DDBPD** [25] is a new direction binary code learning method for palmprint recognition. It extracts informative convolution difference vectors from palmprint patterns, and learn convolution difference vector (CDV) as feature container, which can be used for palmprint recognition.
- **LDDBP** [6] is a direction-based palmprint recognition method, which finds the most discriminant direction features based on exponential and Gaussian fusion model (EGM).
- **PalmNet** [7] applies Gabor filters in CNN to extract discriminative palmprint-specific representation, which is aimed to adapt different images in heterogeneous databases.
- **DEH** [26] trains multiple weak feature extractors based on the online gradient boosting model, and combines them as a single feature extractor. In DEH, activation loss and adversarial loss are adopted to improve the performance.
- **DAN** [40] adopted MK-MMD loss to align the features of multiple task-specific layers.
- **ADDA** [43] is an adversarial learning-based domain adaptation algorithm. It contains a discriminator used to obtain adaptive features. Here, VGG-16-based DHN is also adopted as feature extractor and a discriminator is introduced a feature extractor to obtain the hashing codes in two palmprint datasets.
- **Deep CORAL** [58] is a simple yet effective domain adaptation method. It aligns the second-order statistics of source and target distributions in deep neural networks with a linear transformation.
- **PlamGAN** [14] is proposed for cross-dataset palmprint recognition. It firstly generates fake images based on CycleGAN, and then adopts fake images to train feature extractor used for target dataset. Here, VGG-16-based DHN is also adopted as feature extractor.
- **CycleGAN** [13]. In this paper, CycleGAN is used to generate fake target images and perform cross-dataset palmprint recognition like PalmGAN.
- **CycADA** [48] also performs domain adaptation on pixel and feature levels. After style transferring, ADDA is adopted to obtain alignment features using fake and target images.

Note that for the deep modules above, the backbone adopts the same network architecture. From the results in Tables XI and XII, our JPFA can outperform other models on both cross-dataset palmprint identification and verification. ALDC, DDBPD, and LDDBP are traditional palmprint recognition algorithms and extract the direction information of palmprint according to experience. They can also get relatively good results. From the comparisons, the domain adaptation-based methods may be more suitable for cross-dataset palmprint recognition, and their performances are better, such as PalmGAN and CycADA. However, these methods mainly adopt adversarial learning strategy to obtain adaptive features in the target dataset, but it is difficult to train. JPFA combines MK-MMD loss with consistency loss on feature-level alignment, so it can obtain aligned features more easily. For CycleGAN, there is no identity loss used, and it is even worse than PalmGAN. In addition, compared with one stage-alignment including pixel or feature level, the methods using two stage-alignment can also obtain better results, such as CycADA and our JPFA.

## VI. Conclusion

In this paper, a novel JPFA framework is proposed for cross-dataset palmprint recognition based on two-stage alignment, *i.e.*, pixel-level and feature-level alignment. On the pixel level, fake images are firstly generated by deep style transfer method. The fake images are similar to the target images in style, which can reduce the dataset gaps significantly. In addition, through an identity loss, the identity information is maintained, *i.e.*, the fake images generated by the real images from the same category remains the same category, which can be used for data augmentation. On the feature level, target images are selected to form pairs with source and fake images. After extracting features, MK-MMD loss is adopted to reduce the distribution difference between feature pairs. Further, a novel consistency loss is introduced to constrain the codes extracted by different feature extractors to improve the performance. Finally, adaptive features are obtained to achieve cross-dataset recognition. Multiple experiments are conducted on constrained multispectral databases and unconstrained databases. The results show that our JPFA can effectively improve the performance of cross-dataset palmprint recognition and outperform other models to achieve the state-of-the-arts. Compare with baseline, the accuracy is effectively improved by up to 28.10% and the EER is reduced by up to 4.69%.

## References

[1] A. K. Jain, A. Ross, and S. Prabhakar, "An introduction to biometric recognition," *IEEE Transactions on Circuits and Systems for Video Technology,* Article vol. 14, no. 1, pp. 4-20, Jan 2004,

[2] X. Yin, Y. Zhu, and J. Hu, "Contactless Fingerprint Recognition Based on Global Minutia Topology and Loose Genetic Algorithm," *IEEE Transactions on Information Forensics and Security,* vol. 15, no. 1, pp. 28-41, Jan. 2020,

[3] C. Ding and D. Tao, "Trunk-Branch Ensemble Convolutional Neural Networks for Video-Based Face Recognition," *IEEE Transactions on Pattern Analysis and Machine Intelligence,* Article vol. 40, no. 4, pp. 1002-1014, Apr 2018,

[4] A. Kong, D. Zhang, and M. Kamel, "A survey of palmprint recognition," *Pattern Recognition,* Review vol. 42, no. 7, pp. 1408-1418, Jul 2009,

[5] L. Fei, G. Lu, W. Jia, S. Teng, and D. Zhang, "Feature Extraction Methods for Palmprint Recognition: A Survey and Evaluation," *IEEE Transactions on Systems Man Cybernetics-Systems,* Article vol. 49, no. 2, pp. 346-363, Feb 2019,

[6] L. Fei, B. Zhang, Y. Xu, D. Huang, W. Jia, and J. Wen, "Local Discriminant Direction Binary Pattern for Palmprint Representation and Recognition," *IEEE Transactions on Circuits and Systems for Video Technology,* vol. 30, no. 2, pp. 468-481, 2020,

[7] A. Genovese, V. Piuri, K. N. Plataniotis, and F. Scotti, "PalmNet: Gabor-PCA Convolutional Networks for Touchless Palmprint Recognition," *IEEE Transactions on Information Forensics and Security,* Article vol. 14, no. 12, pp. 3160-3174, Dec 2019,

[8] L. Fei, B. Zhang, W. Zhang, and S. Teng, "Local apparent and latent direction extraction for palmprint recognition," *Information Sciences,* vol. 473, pp. 59-72, 2019,

[9] A.-S. Ungureanu, S. Salahuddin, and P. M. Corcoran, "Towards Unconstrained Palmprint Recognition on Consumer Devices: a Literature Review," *IEEE Access,* pp. 1-19, 2020,

[10] D. Zhong, H. Shao, and X. Du, "A Hand-Based Multi-Biometrics via Deep Hashing Network and Biometric Graph Matching," *IEEE Transactions on Information Forensics and Security,* Article vol. 14, no. 12, pp. 3140-3150, Dec 2019,

[11] S. J. Pan and Q. Yang, "A Survey on Transfer Learning," *IEEE Transactions on Knowledge and Data Engineering,* vol. 22, no. 10, pp. 1345-1359, 2010,

[12] W. M. Matkowski, T. Chai, and A. W. K. Kong, "Palmprint Recognition in Uncontrolled and Uncooperative Environment," *IEEE Transactions on Information Forensics and Security,* vol. 15, pp. 1601 - 1615, 2020,

[13] J. Zhu, T. Park, P. Isola, and A. A. Efros, "Unpaired Image-to-Image Translation using Cycle-Consistent Adversarial Networks," in *16th IEEE International Conference on Computer Vision (ICCV)*, Venice, ITALY, 2017, pp. 2242-2251.

[14] H. Shao, D. Zhong, and Y. Li, "PalmGAN for Cross-domain Palmprint Recognition," in *2019 IEEE International Conference on Multimedia and Expo (ICME)*, Shanghai, China, 2019, pp. 1390-1395.

[15] D. Zhong, X. Du, and K. Zhong, "Decade progress of palmprint recognition: A brief survey," *Neurocomputing,* vol. 328, pp. 16-28, Feb 2019,

[16] P. Dubey, T. Kanumuri, and R. Vyas, "Sequency codes for palmprint recognition," *Signal Image and Video Processing,* Article vol. 12, no. 4, pp. 677-684, May 2018,

[17] X. Wu, D. Zhang, and K. Wang, "Palm line extraction and matching for personal authentication," *IEEE Transactions on Systems Man and Cybernetics Part a-Systems and Humans,* Article vol. 36, no. 5, pp. 978-987, Sep 2006,

[18] W. Jia, D. Huang, and D. Zhang, "Palmprint verification based on robust line orientation code," *Pattern Recognition,* Article vol. 41, no. 5, pp. 1504-1513, May 2008,

[19] Z. Guo, D. Zhang, L. Zhang, and Z. Zuo, Wangmeng, "Palmprint verification using binary orientation co-occurrence vector," *Pattern Recognition Letters,* Article vol. 30, no. 13, pp. 1219-1227, Oct 2009,

[20] L. Zhang, H. Li, and J. Niu, "Fragile Bits in Palmprint Recognition," *IEEE Signal Processing Letters,* Article vol. 19, no. 10, pp. 663-666, Oct 2012,

[21] W. Zuo, Z. Lin, Z. Guo, and D. Zhang, "The Multiscale Competitive Code via Sparse Representation for Palmprint Verification," in *23rd IEEE Conference on Computer Vision and Pattern Recognition (CVPR)*, San Francisco, CA, 2010, pp. 2265-2272.

[22] L. Fei, B. Zhang, Y. Xu, and L. Yan, "Palmprint Recognition Using Neighboring Direction Indicator," *IEEE Transactions on Human-Machine Systems,* Article vol. 46, no. 6, pp. 787-798, Dec 2016,

[23] Y. Xu, L. Fei, J. Wen, and D. Zhang, "Discriminative and Robust Competitive Code for Palmprint Recognition," *IEEE Transactions on Systems Man Cybernetics-Systems,* Article vol. 48, no. 2, pp. 232-241, Feb 2018,

[24] D. Palma, P. L. Montessoro, G. Giordano, and F. Blanchini, "Biometric Palmprint Verification: A Dynamical System Approach," *IEEE Trans. Systems, Man, and Cybernetics: Systems,* vol. 49, no. 12, pp. 2676-2687, 2019,

[25] L. Fei, B. Zhang, Y. Xu, Z. Guo, J. Wen, and W. Jia, "Learning Discriminant Direction Binary Palmprint Descriptor," *IEEE Transactions on Image Processing,* Article vol. 28, no. 8, pp. 3808-3820, Aug 2019,

[26] H. Shao, D. Zhong, and X. Du, "Effective deep ensemble hashing for open-set palmprint recognition," *Journal of Electronic Imaging,* vol. 29, no. 01, pp. 1-19, 2020,

[27] S. Chen, S. Chen, Z. Guo, and Y. Zuo, "Low-resolution palmprint image denoising by generative adversarial networks," *Neurocomputing,* Article vol. 358, pp. 275-284, Sep 2019,

[28] W. Jia, R. Hu, J. Gui, Y. Zhao, and X. Ren, "Palmprint Recognition across Different Devices," *Sensors,* vol. 12, no. 6, pp. 7938-7964, 2012,

[29] A. S. Ungureanu, S. Thavalengal, T. E. Cognard, C. Costache, and P. Corcoran, "Unconstrained Palmprint as a Smartphone Biometric," *IEEE Transactions on Consumer Electronics,* Article vol. 63, no. 3, pp. 334-342, Aug 2017,

[30] C. Tan, F. Sun, T. Kong, W. Zhang, C. Yang, and C. Liu, "A Survey on Deep Transfer Learning," in *27th International Conference on Artificial Neural Networks (ICANN)*, Rhodes, GREECE, 2018, vol. 11141, pp. 270-279.
[31] K. Weiss, T. M. Khoshgoftaar, and D. Wang, "A survey of transfer learning," *Journal of Big Data,* vol. 3, no. 1, pp. 1-40, Dec. 2016,
[32] I. J. Goodfellow *et al.*, "Generative Adversarial Nets," in *2014 Advances in Neural Information Processing Systems*, Montreal, Quebec, Canada, 2014, pp. 2672-2680.
[33] P. Isola, J. Zhu, T. Zhou, and A. A. Efros, "Image-to-Image Translation with Conditional Adversarial Networks," presented at the 30th IEEE Conference on Computer Vision and Pattern Recognition, Hawaii, USA, 2017.
[34] L. Wei, S. Zhang, W. Gao, and Q. Tian, "Person Transfer GAN to Bridge Domain Gap for Person Re-Identification," in *31st IEEE/CVF Conference on Computer Vision and Pattern Recognition (CVPR)*, Salt Lake City, UT, 2018, pp. 79-88.
[35] W. Deng, L. Zheng, Q. Ye, G. Kang, Y. Yang, and J. Jiao, "Image-Image Domain Adaptation with Preserved Self-Similarity and Domain-Dissimilarity for Person Re-identification," in *31st IEEE/CVF Conference on Computer Vision and Pattern Recognition (CVPR)*, Salt Lake City, UT, 2018, pp. 994-1003.
[36] Z. Zhong, L. Zheng, Z. Zheng, S. Li, and Y. Yang, "Camera Style Adaptation for Person Re-identification," in *31st IEEE/CVF Conference on Computer Vision and Pattern Recognition (CVPR)*, Salt Lake City, USA, 2018, pp. 5157-5166.
[37] H. Dou, C. Chen, X. Hu, and S. Peng, "Asymmetric CycleGAN for unpaired NIR-to-RGB face image translation," in *44th IEEE International Conference on Acoustics, Speech and Signal Processing (ICASSP)*, Brighton, ENGLAND, 2019, pp. 1757-1761.
[38] S. Pan, I. W. Tsang, J. T. Kwok, and Q. Yang, "Domain Adaptation via Transfer Component Analysis," *IEEE Transactions on Neural Networks,* Article vol. 22, no. 2, pp. 199-210, Feb 2011,
[39] M. Long, J. Wang, G. Ding, J. Sun, and P. Yu, "Transfer Feature Learning with Joint Distribution Adaptation," in *IEEE International Conference on Computer Vision (ICCV)*, Sydney, AUSTRALIA, 2013, pp. 2200-2207.
[40] M. Long, Y. Cao, J. Wang, and M. Jordan, "Learning Transferable Features with Deep Adaptation Networks," in *32nd International Conference on Machine Learning*, Lille, France, 2015, pp. 97-105.
[41] Y. Li, N. Wang, J. P. Shi, X. Hou, and J. Liu, "Adaptive Batch Normalization for practical domain adaptation," *Pattern Recognition,* Article vol. 80, pp. 109-117, Aug 2018,
[42] Y. Ganin *et al.*, "Domain-Adversarial Training of Neural Networks," *Journal of Machine Learning Research,* Article vol. 17, pp. 1-35, 2016,
[43] E. Tzeng, J. Hoffman, K. Saenko, and T. Darrell, "Adversarial Discriminative Domain Adaptation," in *30th IEEE/CVF Conference on Computer Vision and Pattern Recognition (CVPR)*, Honolulu, HI, 2017, pp. 2962-2971.
[44] C. Chen, Z. Chen, B. Jiang, and X. Jin, "Joint Domain Alignment and Discriminative Feature Learning for Unsupervised Deep Domain Adaptation," in *33th AAAI Conference on Artificial Intelligence (AAAI)*, Honolulu, Hawaii, USA, 2019, pp. 3296-3303.
[45] C. Chen, Q. Dou, H. Chen, and P. Heng, "Semantic-Aware Generative Adversarial Nets for Unsupervised Domain Adaptation in Chest X-Ray Segmentation," in *9th International Workshop on Machine Learning in Medical Imaging (MLMI)*, Granada, SPAIN, 2018, vol. 11046, pp. 143-151.
[46] H. Shao, D. Zhong, and X. Du, "Cross-Domain Palmprint Recognition Based on Transfer Convolutional Autoencoder," in *2019 IEEE International Conference on Image Processing (ICIP)*, Taipei, Taiwan, 2019, pp. 1153-1157.
[47] Y. Chen, Y. Lin, M. Yang, and J. Huang, "CrDoCo: Pixel-Level Domain Transfer With Cross-Domain Consistency," in *IEEE Conference on Computer Vision and Pattern Recognition (CVPR)*, Long Beach, CA, USA, 2019, pp. 1791-1800.
[48] J. Hoffman *et al.*, "CyCADA: Cycle-Consistent Adversarial Domain Adaptation," in *35th International Conference on Machine Learning (ICML)*, Stockholm, Sweden, 2018, pp. 1994-2003.
[49] H. Liu, R. Wang, S. Shan, and X. Chen, "Deep Supervised Hashing for Fast Image Retrieval," in *2016 IEEE Conference on Computer Vision and Pattern Recognition (CVPR)*, Seattle, WA, 2016, pp. 2064-2072.
[50] W. Deng, L. Zheng, Q. Ye, G. Kang, Y. Yang, and J. Jiao, "Image-Image Domain Adaptation with Preserved Self-Similarity and Domain-Dissimilarity for Person Re-identification," in *IEEE Conference on Computer Vision and Pattern Recognition*, Salt Lake City, USA, 2018, pp. 1-10.
[51] A. Gretton, K. M. Borgwardt, M. J. Rasch, B. Schölkopf, and A. J. Smola, "A Kernel Two-Sample Test," *The Journal of Machine Learning Research,* vol. 13, pp. 723-773, 2012,
[52] H. Shao, D. Zhong, and X. Du, "Towards Efficient Unconstrained Palmprint Recognition via Deep Distillation Hashing," *arXiv,* pp. 1-13, 2020,
[53] H. Shao, D. Zhong, and X. Du, "Efficient Deep Palmprint Recognition via Distilled Hashing Coding," in *Conference on Computer Vision and Pattern Recognition Workshops*, Long Beach, CA, USA, 2019, pp. 714-723.
[54] D. Zhang, Z. Guo, G. Lu, L. Zhang, and W. Zuo, "An Online System of Multispectral Palmprint Verification," *IEEE Transactions on Instrumentation and Measurement,* Article vol. 59, no. 2, pp. 480-490, Feb 2010,
[55] Y. Zhang, L. Zhang, R. Zhang, S. Li, J. Li, and F. Huang, "Towards Palmprint Verification On Smartphones," *arXiv,* pp. 1-14, 2020,
[56] K. Simonyan and A. Zisserman, "Very Deep Convolutional Networks for Large-Scale Image Recognition," in *3rd International Conference on Learning Representations (ICLR)*, San Diego, CA, USA, 2015.
[57] J. Deng, W. Dong, R. Socher, L. Li, K. Li, and F. Li, "ImageNet: A large-scale hierarchical image database," in *Conference on Computer Vision and Pattern Recognition (CVPR)*, Miami, Florida, 2009, pp. 248-255.
[58] B. Sun and K. Saenko, "Deep CORAL: Correlation Alignment for Deep Domain Adaptation," in *European Conference on Computer Vision Workshops*, Amsterdam, The Netherlands, 2016, pp. 443-450.